\documentclass[10pt,twocolumn]{article}

\usepackage[letterpaper,margin=1in]{geometry}
\usepackage[T1]{fontenc}
\usepackage[utf8]{inputenc}
\usepackage{times}
\usepackage{inconsolata}
\usepackage{microtype}
\usepackage{amsmath}
\usepackage{amssymb}
\usepackage{amsfonts}
\usepackage{booktabs}
\usepackage{graphicx}
\graphicspath{{figures/}}
\usepackage[hyphens]{url}
\usepackage[hidelinks]{hyperref}
\usepackage{natbib}

\hypersetup{
  pdftitle={Is It Still Worth Training a Classical Model in the Era of LLMs? A Crossover Benchmark on Tabular Data},
  pdfauthor={Kaihua Ding}}

\begin{document}

\twocolumn[{%
  \centering
  {\LARGE\bfseries Is It Still Worth Training a Classical Model\\[2pt]
   in the Era of LLMs?\par}
  \vspace{5pt}
  {\Large A Crossover Benchmark on Tabular Data\par}
  \vspace{15pt}
  {\large Kaihua Ding\par}
  \vspace{4pt}
  {University of Pennsylvania\par}
  \vspace{2pt}
  {\texttt{dkaihua@upenn.edu}\par}
  \vspace{17pt}
  \begin{minipage}{0.86\textwidth}
    \begin{center}\textbf{Abstract}\end{center}
    \small\noindent Large language models can label a tabular row from a plain-English description
with no training---a capability now shipping in mainstream spreadsheet tools such
as Microsoft Copilot in Excel and Anthropic's Claude for Excel
\citep{microsoft2025copilotexcel, anthropic2025claudeexcel}---raising a
practical question for the many business prediction
problems where labels are expensive: should you prompt a frozen LLM, or collect
data and train a model---and if so, how much data? We quantify the answer with
the \emph{labeled-data crossover} $N^\star$, the training-set size at which a
trained classical model's learning curve overtakes a frozen LLM's
training-free (and therefore flat) error. Aggregating 126 independent student
evaluations of small GPT models under eight prompting configurations across 18
tabular datasets, paired with authoritative power-law learning curves for six
classical model families, we find that training wins fast: even given an oracle
choice of its best prompt configuration, a trained classical model beats the
small frozen LLM using no more labeled data than is already on hand in 86\% of
cases, and wins by the smallest labeled subset we evaluate in 40\%, with the
observed crossover at a median of $\sim$6\% of the training set. In-context
few-shot examples do not behave like training---error versus shot count does not
follow a power law---and the same protocol re-run by independent implementers
varies with a coefficient of variation of $0.148$. A controlled probe indicates the LLM
depends on recognizable feature-name semantics, which plausibly makes our
crossover a conservative estimate (we do not claim memorization). For a typical business
table, the evidence is clear: collect a few hundred labels and train a
gradient-boosted model.

  \end{minipage}
  \vspace{18pt}
}]

\section{Introduction}
\label{sec:introduction}

Large language models can be handed a row of a spreadsheet, described in plain
English, and asked to predict its label---no training, no feature engineering,
no model selection \citep{hegselmann2023tabllm, dinh2022lift, gardner2024tabula}.
This is no longer a research curiosity: the same capability now ships inside the
tools practitioners use every day---from Microsoft Copilot in Excel and
Anthropic's Claude for Excel to ChatGPT's data-analysis mode and Gemini in Google
Sheets---each letting an analyst point a frozen LLM at a spreadsheet column and
ask for a prediction \citep{microsoft2025copilotexcel, anthropic2025claudeexcel,
openai2024chatgptdata, google2025geminisheets}. This is appealing for the long tail of business prediction
problems (churn, default risk, conversion, pricing) where collecting and
labeling data is the dominant cost. It raises a concrete deployment question
that the practitioner actually faces: \emph{should I prompt a frozen LLM, or
collect labels and train a model---and if the latter, how many labels do I
need?}

We answer this with a quantity we call the \textbf{labeled-data crossover}
$N^\star$: the training-set size at which a trained classical model's learning
curve drops below a frozen LLM's (training-free, and therefore flat) error. To
the left of $N^\star$ the LLM is the rational choice; to the right, training
wins. Mapping $N^\star$ across many datasets, model families, and prompting
choices turns the vague ``LLMs are good at tables now'' discourse into an
actionable decision boundary.

We estimate this boundary at a scale rarely available for LLM evaluation. In a
graduate course, 126 students each independently evaluated three of 18 tabular
business datasets under eight prompting configurations ($2$ models $\times$ $2$
serializations $\times$ $2$ shot settings), on a fixed test set of up to 100 rows
per dataset---roughly 3{,}000 evaluation cells, each replicated by 16--29
independent implementations. We pair these LLM measurements with authoritative
power-law learning curves for six classical model families, fit on the same
datasets in a companion study, and recompute every crossover from a single
source of truth.

Our findings are consistently \emph{cautionary} for the ``just prompt an LLM''
position:
\begin{itemize}
  \item \textbf{Training wins, and fast.} Even against an oracle choice of each
  dataset's \emph{best} LLM configuration, a trained classical model beats the
  frozen LLM using \textbf{no more labeled data than is already on hand in 86\%}
  of (dataset, model) cells (70 reliable classical curves; 10 unreliable excluded),
  and wins by the smallest labeled subset we evaluate in \textbf{40\%}; where the
  crossover falls within the observed data its median is only \textbf{$\sim$6\%}.
  Gradient boosting overtakes the LLM on every dataset but one synthetic table; the
  cells it never wins are weak baselines (\S\ref{sec:crossover}).
  \item \textbf{In-context examples do not rescue it.} Few-shot prompting does not
  behave like training: error-versus-shots follows no power law (only 10\% of fits
  reach $R^2>0.8$; median exponent $\approx$0) (\S\ref{sec:config}).
  \item \textbf{Decisions are noisy under replication.} The same fixed protocol
  yields a median coefficient of variation of $0.148$ in LLM error across
  independent re-implementations, which propagates into the crossover
  (\S\ref{sec:variance}).
  \item \textbf{The skill is value-dependent.} A factorial probe with a trained
  positive control shows the LLM's edge collapses when exact values are replaced by
  readable deciles---though a model trained on the same bins keeps its accuracy---while
  anonymizing names has weaker, inconsistent effects. The LLM leans on recognizable,
  exact-valued inputs, which \emph{suggests}---without proving---that our crossover
  is conservative (\S\ref{sec:contamination}).
\end{itemize}

Taken together, for the realistic case of a small, inexpensive LLM on a typical
business table, the evidence says: collect a few hundred labels and train a
gradient-boosted model. We release the aggregated curves, the canonical
crossover table, and all code to support practitioners making this call.

\section{Related Work}
\label{sec:related}

\paragraph{LLMs as tabular predictors.}
A growing line of work serializes tabular rows into text and applies language
models. TabLLM \citep{hegselmann2023tabllm} shows that a fine-tuned LLM is a
strong \emph{few-shot} tabular classifier; LIFT \citep{dinh2022lift} fine-tunes
LLMs for general non-language tasks via natural-language interfaces;
\citet{manikandan2023weak} use frozen LLMs as weak learners inside boosting; and
TabuLa-8B \citep{gardner2024tabula} pretrains a tabular foundation model on
large corpora of tables. These efforts build \emph{methods} that make LLMs
better at tables. We instead ask a deployment question that none of them answers
directly: for an off-the-shelf, \emph{frozen} model and a realistic business
table, how much labeled data must a practitioner collect before simply training
a classical model wins? We also focus on small, cheap production models
(nano/mini) rather than frontier or fine-tuned systems.

\paragraph{Small-data tabular learning.}
TabPFN \citep{hollmann2023tabpfn,hollmann2025tabpfn} is a prior-fitted
transformer that excels in the very-low-data regime. It is complementary to our
study: it is a purpose-built tabular model rather than a general LLM prompted in
natural language, and it sharpens the practical question of \emph{which}
training-free option to reach for at small $N$. A parallel line builds
specialized deep tabular architectures
\citep{gorishniy2021revisiting,arik2021tabnet,popov2020neural,kadra2021welltuned} and
surveys the space \citep{borisov2022deep}; we use standard trained baselines
rather than these, as they are what a practitioner deploys by default.

\paragraph{Tree ensembles still dominate tabular data.}
Gradient-boosted trees \citep{chen2016xgboost} and random forests continue to
outperform deep learning on typical tabular problems
\citep{grinsztajn2022why,shwartzziv2022tabular}, and even careful studies of
\emph{when} neural nets win still find tree ensembles the safe default
\citep{mcelfresh2023neural}. These are exactly the trained baselines our
crossover measures the frozen LLM against; AutoGluon-Tabular
\citep{erickson2020autogluon} packages them into one AutoML system.

\paragraph{Memorization and contamination.}
LLMs are known to memorize pretraining data verbatim
\citep{carlini2021extracting,carlini2023quantifying} and can exploit test data
leaked into pretraining \citep{magar2022data}, and \citet{bordt2024elephants} show this
specifically for popular tabular datasets, confounding any claim that an LLM
``predicts'' such data. This motivates our contamination probe
(\S\ref{sec:contamination}): we treat pretraining exposure as a measured axis.
We find the frozen LLM leans on recognizable feature semantics, which---whether
from memorization or general world knowledge---can only \emph{help} it on these
familiar datasets, making our crossover a conservative, LLM-favorable estimate.

\paragraph{Scaling and learning curves.}
Neural scaling laws predict performance from compute, data, and model size
\citep{kaplan2020scaling,hoffmann2022training,hestness2017deep}, while a classical
learning-curve literature fits power laws to tree, kernel, and linear models
\citep{perlich2003tree,mukherjee2003estimating,figueroa2012predicting,domhan2015speeding}
(see \citealp{viering2023shape} for a review).
We connect these regimes to the training-free LLM regime through a single
decision-relevant quantity: the labeled-data crossover.

\section{Experimental Setup}
\label{sec:setup}

\paragraph{Datasets.}
We study 18 tabular business datasets drawn from standard repositories
(UCI, Kaggle, and library built-ins): 10 classification tasks
(\texttt{adult\_income}, \texttt{bank\_marketing}, \texttt{credit\_card\_default},
\texttt{diabetes\_pima}, \texttt{employee\_attrition}, \texttt{german\_credit},
\texttt{heart\_disease}, \texttt{online\_shoppers}, \texttt{telco\_churn},
\texttt{wine\_quality}) and 8 regression tasks (\texttt{abalone},
\texttt{ames\_housing}, \texttt{auto\_mpg}, \texttt{bike\_sharing},
\texttt{california\_housing}, \texttt{concrete\_strength}, \texttt{diamonds},
\texttt{insurance\_charges}). They span domains (finance, HR, marketing,
healthcare, real estate, e-commerce) and sizes from 303 to 53{,}940 rows.
These are precisely the kind of small-to-medium business tables where a
practitioner must decide between collecting labels to train a model and simply
prompting an off-the-shelf LLM.

\paragraph{Distributed replication protocol.}
The measurements come from a graduate machine-learning course in which 126
students each independently evaluated their three assigned datasets following a
fixed starter notebook (\texttt{random\_state}=42, 80/20 train/test split, a
fixed test set of up to 100 rows per dataset---fewer for the smallest datasets,
e.g.\ 61 for Heart Disease and 80 for Auto MPG). Because dataset assignments overlap, every
(dataset, configuration) cell is replicated by 16--29 independent
implementations (median 18). This yields both pooled point estimates and a
direct measure of cross-replicator variance under a nominally identical
protocol (\S\ref{sec:variance}), at a replication scale rarely available for
LLM evaluations.

\paragraph{LLM configurations.}
Each dataset is evaluated under $2\times2\times2=8$ configurations: two models
(\texttt{gpt-4.1-nano}, \texttt{gpt-4.1-mini}), two serializations of a row into
a prompt (\emph{key--value}, e.g.\ ``\texttt{age: 67, duration: 6, \dots}'', vs.\
\emph{natural language} generated from feature descriptions), and two in-context
settings (zero-shot vs.\ 5-shot, with exemplars drawn only from the training
split). Each configuration is scored on the same fixed test rows (up to 100), giving up
to 800 API calls for each student's assigned dataset. We deliberately study small, inexpensive production models
rather than frontier or fine-tuned ones: they are the realistic default a
cost-conscious practitioner would reach for. We additionally analyze two
optional student extensions: a few-shot sweep $k\in\{1,3,5,10,20\}$
(\S\ref{sec:config}) and optional \texttt{gpt-4.1} (bigger-model) runs that we pair
within-student against each student's own nano/mini (a suggestive bonus check).

\paragraph{Metrics.}
For classification the primary metric is AUROC and we report
$\mathrm{error}=1-\mathrm{AUROC}$; for regression we report the normalized error
$\mathrm{RMSE}/\mathrm{std}(y_{\text{test}})$. Both put $0$ at perfect prediction,
but their trivial-baseline levels differ: chance ranking gives classification
error $0.5$, whereas a mean predictor gives regression error $1.0$. We therefore
never compare absolute error \emph{magnitudes} across task types; each crossover
is computed within a single dataset's metric, where the LLM and classical errors
are directly comparable.

\paragraph{Classical baselines and the crossover.}
The classical side reuses a companion study in which 127 students fit power-law
learning curves $\mathrm{error}(N)=aN^{-b}+c$ for six model families across the
same 18 datasets; we take the per-(dataset, model) \emph{pooled} fits as
authoritative. Classification uses
\{Boosting, RandomForest, SVM, LinearModel\} and regression uses
\{Boosting, RandomForest, SVM, Ridge, Lasso\}. Because an LLM is evaluated
without task-specific training, its error is a horizontal line independent of
$N$. For each (dataset, classical model, LLM configuration) we solve for the
\emph{crossover} training-set size at which the classical curve meets the LLM:
\begin{equation}
N^\star=\left(\frac{a}{e_{\text{LLM}}-c}\right)^{1/b},
\label{eq:crossover}
\end{equation}
where $e_{\text{LLM}}$ is the median LLM error across replicators. If
$e_{\text{LLM}}\le c$ the classical floor never reaches the LLM
(\emph{classical-never-better}); if $N^\star$ falls below the smallest training
size in the classical curves ($N_{\min}$, the $\sim$1\% data fraction, ranging
from a few rows to a few hundred across datasets) the classical model already
wins (\emph{classical-always-better}). We report $N^\star$ both in absolute rows and
as a fraction $N^\star/N_{\text{full}}$ of the available training set.

\paragraph{Fairness of the comparison.}
The classical side is a \emph{pooled} fit across 127 students and the LLM side the
\emph{median} over implementations: we pool because the deployment question
concerns a standard trained model, not one noisy run, and the LLM is held to an
equally fixed protocol. We check that pooling does not drive the result by
comparing against per-student crossovers (\S\ref{sec:variance}) and report the
conservative oracle-best-config view as the headline.

\section{The Labeled-Data Crossover}
\label{sec:crossover}

\begin{figure*}[t]
\centering
\includegraphics[width=\textwidth]{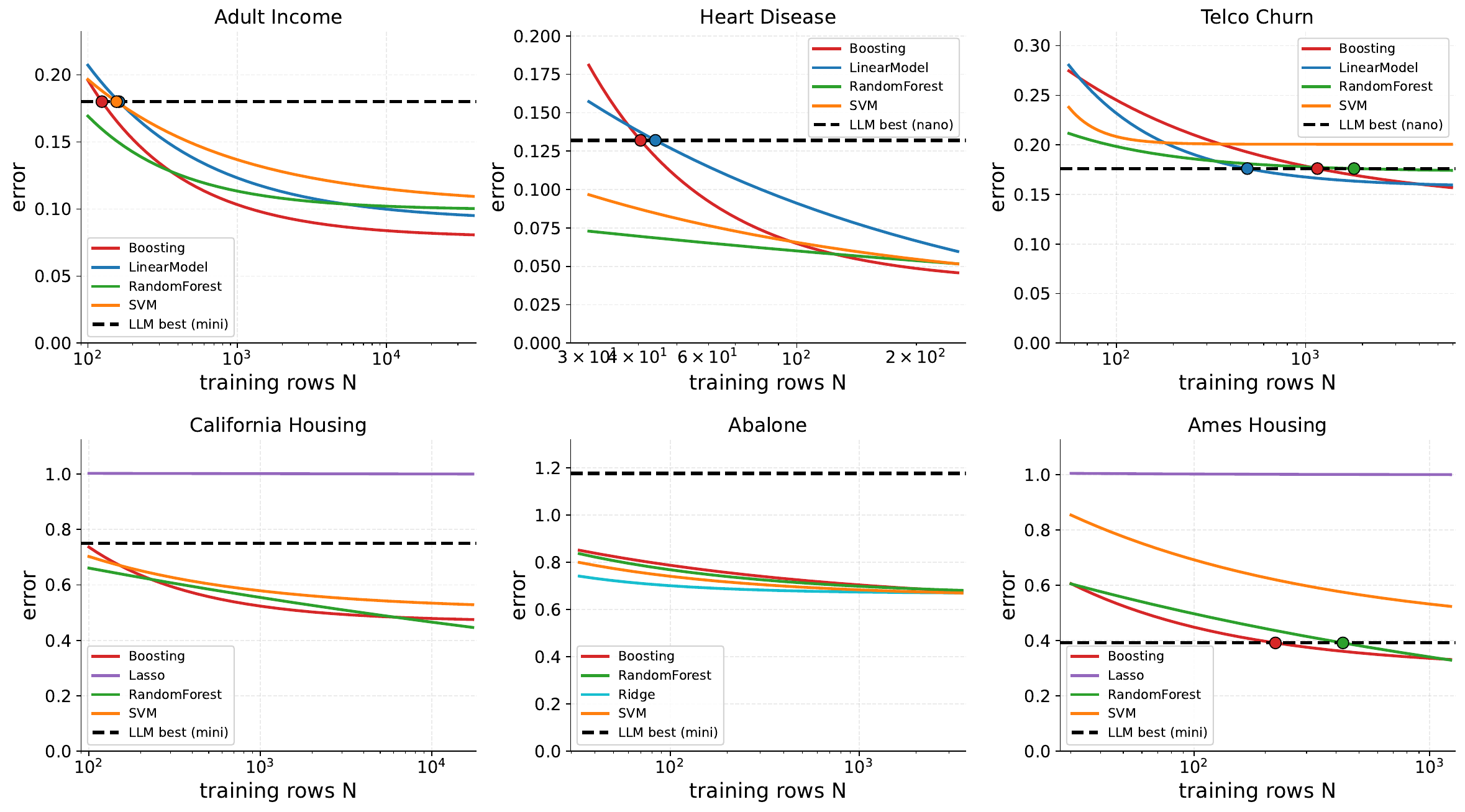}
\caption{Classical learning curves $\mathrm{error}(N)=aN^{-b}+c$ (solid) versus
the best frozen-LLM configuration's training-free error (dashed horizontal) on
six representative datasets. The marker shows the crossover $N^\star$. On most
(dataset, model) pairs the classical curve drops below the LLM within a small
fraction of the available training data.}
\label{fig:hero}
\end{figure*}

Our central question is how much labeled data a practitioner must collect before
a trained classical model beats a frozen LLM. We answer it with the crossover
$N^\star$ of Eq.~\ref{eq:crossover}, comparing each dataset's pooled classical
curves against the LLM. To avoid flattering the classical side, we report the
\emph{conservative} comparison: each dataset against its single best-performing
LLM configuration---the one of eight with the lowest median error. This ``best''
is an \emph{oracle}, chosen post hoc on the evaluation data and therefore not a
deployable selection rule; it deliberately upper-bounds the frozen LLM, and the
crossover is small even so.

\paragraph{How often, and after how much data, does training win?}
We split each reliable (dataset, classical-model) cell---70 of the 80 nominal
dataset$\times$model curves pass a pre-specified reliability filter ($R^2>0.5$,
$0<b<3$, converged); the 10 excluded are near-flat or non-converged classical
fits---into four decision-relevant cases: the classical model already wins at the smallest
training size we observe ($N^\star\!\le\!N_{\min}$), it crosses within the
observed data range ($N_{\min}\!<\!N^\star\!\le\!N_{\text{full}}$), it would
cross only beyond the full training set ($N^\star\!>\!N_{\text{full}}$), or it
never catches the LLM's error floor; cells whose extrapolated $N^\star$ is
astronomically large are kept in the denominator (they count as ``not beaten
within the data''). Even against the oracle LLM configuration, a trained model
beats it using \emph{no more than the labels already on hand} in \textbf{86\%} of
cells (dataset-cluster bootstrap 95\% CI $[76,94]$), and wins by the
\emph{smallest labeled subset we evaluate} in \textbf{40\%} ($[22,57]$). Where the
crossover falls inside the observed range, it sits at a median of \textbf{5.9\%}
of the full training set ($95\%$ CI $[0.5,17]$; Figure~\ref{fig:hero}). In the
remaining 14\% of cells the LLM survives on the available data---6 where it beats
the classical error floor outright and 4 where the projected crossover lies
beyond the full training set---concentrated in weak baselines (below). The split
is similar across tasks (classification $87\%$ within-data, median $5.3\%$;
regression $84\%$, $6.4\%$), and starker against the LLM's \emph{typical}
configuration (training wins within available data in $95\%$ of cells). The LLM
is not a free win even at $N{=}0$: against its oracle configuration it fails to
beat a trivial majority-class/mean predictor on $1/18$ datasets (Abalone), and
its \emph{typical} small-model regression error ($1.03$) is worse than that of a
constant mean predictor.

\begin{figure*}[t]
\centering
\includegraphics[width=\textwidth]{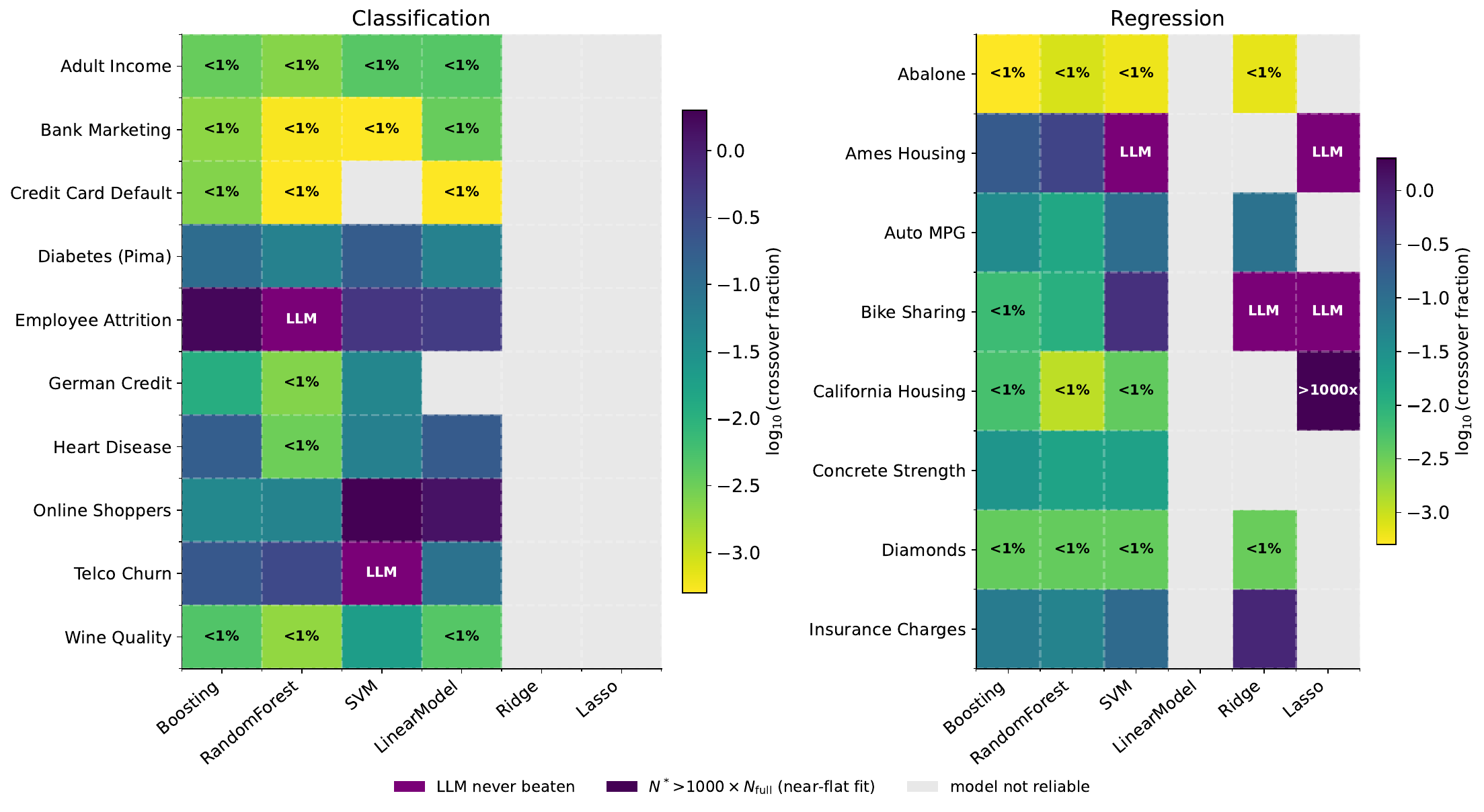}
\caption{Crossover fraction $N^\star/N_\mathrm{full}$ (log scale) for every
(dataset, classical model) against the best LLM configuration. ``$<$1\%'' marks
cells the classical model already wins at the smallest tested size;
``LLM'' marks the few cells (weak baselines) it never catches.}
\label{fig:summary}
\end{figure*}

\paragraph{Strong baselines settle it; only weak ones leave the LLM ahead.}
The crossover depends sharply on the classical model
(Figure~\ref{fig:summary}). Gradient boosting and random forests are \emph{never}
asymptotically unbeatable (0 and 1 never-catch cells), and restricting to these
two strong families, a trained model wins within the available data in 94\% of
cells (median crossover 5.2\%). The lone exception for boosting is the
\emph{synthetic} Employee Attrition data, where its projected crossover lands
just beyond the full training set ($N^\star\!\approx\!1.7N_{\text{full}}$), so the
LLM is not overtaken there within the available labels---a caveat we state rather
than paper over. The six cells where the classical model never catches the LLM
all belong to weak or ill-suited baselines: two are Lasso (which fails to fit
several curves), the rest Ridge, SVM, or a random forest on synthetic Employee
Attrition---baseline weakness, not LLM strength. Slower-fitting linear and kernel
models (LinearModel, SVM, Ridge) yield the largest crossovers (family medians
$4.5$--$5.3\%$), but even these remain a small fraction of the available data.

\paragraph{Robustness and uncertainty.}
The verdict is stable. Sweeping the fit-reliability filter ($R^2>0.3,0.5,0.7$)
moves the immediate-win share only within $40$--$43\%$ and the observed-crossover
median within $5.9$--$6.4\%$, and no retained cell lacks a finite $N^\star$. The
main threat---the LLM is scored on an up-to-100-row \emph{sample}, the classical
curves on the full test split---we address in two parts. (i)~Row-bootstrapping the
samples puts the test-sample SD of $1-\mathrm{AUROC}$ at a median $0.048$;
propagating it as a per-dataset Gaussian SE ($0.05$ clf, $0.08\,e_{\text{LLM}}$
reg) widens the within-available-data interval only to $[70,95]\%$. (ii)~the
regression-normalization correction (\S\ref{sec:limitations}) moves the
observed-crossover median only $5.9\%\!\to\!5.2\%$. (iii)~Bootstrapping the
per-student curve fits to propagate curve-parameter uncertainty into $N^\star$
leaves $67$ of $70$ cells' category unchanged and, stacked with the dataset and
test-sample resamplings, gives a within-available-data interval of $[68,93]\%$. A
model-free check corroborates it: on the \emph{raw} learning-curve points, the
classical model beats the best LLM at the smallest \emph{observed} size in $48\%$
of cells, near the $40\%$ from the fitted curves. Finally, re-scoring each
dataset's best configuration on the full matched test split (up to $2{,}500$ rows)
rather than the $100$-row sample---the one mismatch these resamplings leave
open---and recomputing the crossover leaves the verdict intact ($94\%$ within the
available data, $63\%$ at the smallest size): the LLM's full-split error is if
anything slightly \emph{higher}, so the $100$-row protocol was mildly LLM-favorable
and does not inflate the crossover.

\section{Effect of Configuration}
\label{sec:config}

\begin{figure*}[t]
\centering
\includegraphics[width=\textwidth]{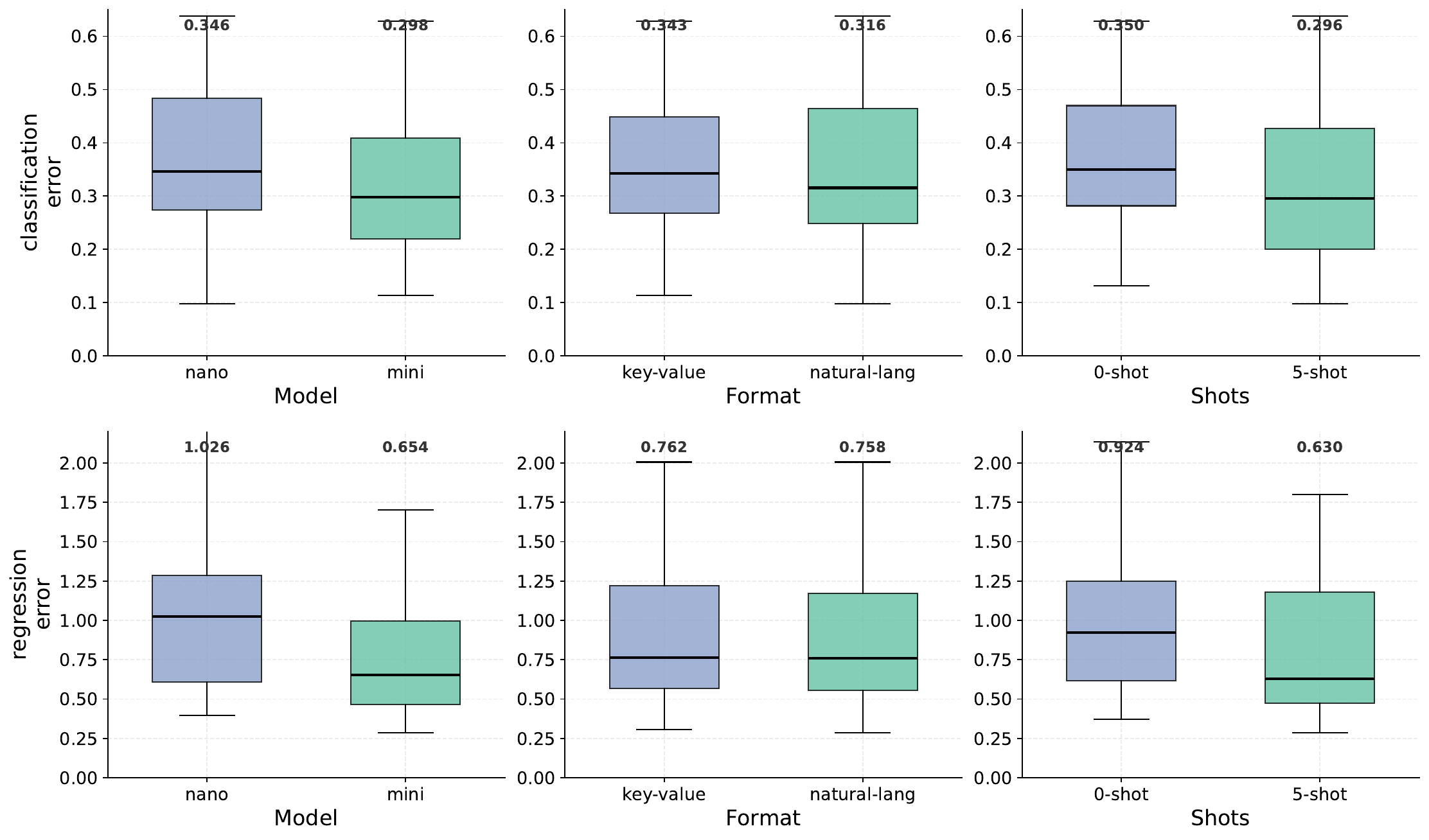}
\caption{Per-cell LLM error along the three configuration axes, split by task
(top: classification, $1-\mathrm{AUROC}$; bottom: regression, normalized RMSE).
Box annotations are medians. A more capable model (\texttt{mini}) and in-context
examples (5-shot) both lower error, most strongly on regression; key--value
vs.\ natural-language serialization is indistinguishable.}
\label{fig:config}
\end{figure*}

Would prompting harder move the crossover? We decompose the eight configurations
along their three binary axes---model capacity, serialization format, and shot
count---with paired contrasts that hold the replicator and the other two axes
fixed (Figure~\ref{fig:config}), and we test separately whether in-context
examples scale like training data.

\paragraph{A more capable model helps; format does not.}
The larger \texttt{mini} model is consistently better than \texttt{nano}---median
error \textbf{0.457} vs.\ \textbf{0.542}---and the gap holds paired within every
matched (dataset, replicator, prompt) cell. It is modest on classification
($+0.033$, \texttt{nano} worse in $70.6\%$ of pairs) but large on regression
($+0.182$, worse in $94.1\%$), where the smaller model frequently emits
near-useless numeric predictions---\texttt{nano}'s typical regression error
($1.03$) in fact exceeds that of a trivial mean predictor (normalized error
$1.0$), a bar \texttt{mini} ($0.65$) clears. Serialization format, in contrast, barely
matters: natural-language vs.\ key--value differs by a paired median of just
$-0.003$ overall ($0.000$ classification, $-0.006$ regression), and the two
distributions in Figure~\ref{fig:config} are nearly identical.

\begin{figure}[t]
\centering
\includegraphics[width=\columnwidth]{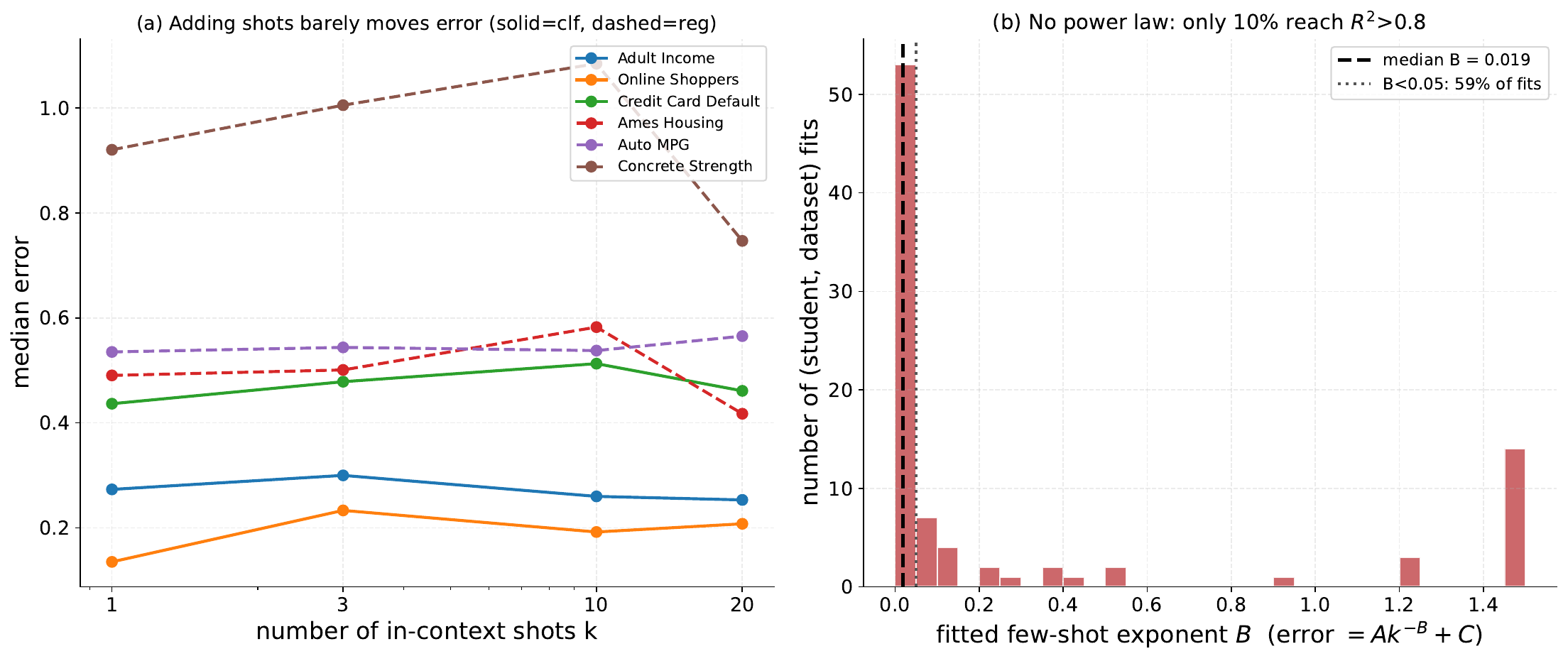}
\caption{In-context few-shot does not scale like training. (a) Median error vs.\
number of shots $k$ for six datasets (solid: classification; dashed: regression):
the curves are nearly flat and occasionally rise. (b) Distribution of the fitted
exponent $B$ in $\mathrm{error}=Ak^{-B}+C$ over (student, dataset) curves; the
median is $0.019$ and $59\%$ of fits have $B<0.05$, against a classical
\emph{training} exponent $b\approx0.5$.}
\label{fig:fewshot}
\end{figure}

\paragraph{In-context examples help once, then stop.}
Moving from zero- to 5-shot lowers error by a paired median of $-0.062$ overall
($-0.044$ classification, $-0.105$ regression): a real but essentially one-time
gain. It does not compound. On a $k\in\{1,3,5,10,20\}$ sweep contributed by $30$
students (of $36$ who submitted bonus runs; $90$ curves, each fit over all five
shot counts) we fit a power law $\mathrm{error}=Ak^{-B}+C$ to each (student,
dataset) curve, and it does not hold---the median fitted exponent is only $B=\mathbf{0.019}$ (against a
classical \emph{training} exponent $b\approx0.5$), just \textbf{10\%} of fits
reach $R^2>0.8$, and \textbf{59\%} have $B<0.05$, i.e.\ are effectively flat
(Figure~\ref{fig:fewshot}b). Comparing each dataset's few-shot exponent against
its classical learning-curve exponent, in-context examples improve more slowly
than training on $15$ of $18$ datasets, and on several (e.g.\ Concrete Strength)
more shots make error \emph{worse} (Figure~\ref{fig:fewshot}a). Numerical
curve-fitting converges on these flat curves too, so a high convergence rate is
not evidence of a law; the near-zero exponent and poor $R^2$ are. The flat LLM
line is a ceiling, not a curve that more context steadily bends down.

\paragraph{Parsing is reliable; serving is not free.}
Output parsing is a non-issue: mean parse rates are $0.998$--$0.999$ across all
eight configurations; the lowest single-dataset rate ($0.73$) occurs once, for
\texttt{nano}/key--value/zero-shot. Cost is the real lever. A median
$100$-row evaluation run costs \$0.005 with \texttt{nano} and \$0.020 with
\texttt{mini}---roughly \$0.00005 and \$0.0002 per row---and
the single most accurate configuration (\texttt{mini}/key--value/5-shot, median
error \textbf{0.419}) costs \$0.031 per run. Every prediction re-incurs this
charge, whereas a trained gradient-boosted model's marginal inference cost is
effectively zero, so the configuration choices that buy accuracy also widen the
LLM's standing cost disadvantage (\S\ref{sec:discussion}).

\section{Variance and Stability}
\label{sec:variance}

\begin{figure}[t]
\centering
\includegraphics[width=\columnwidth]{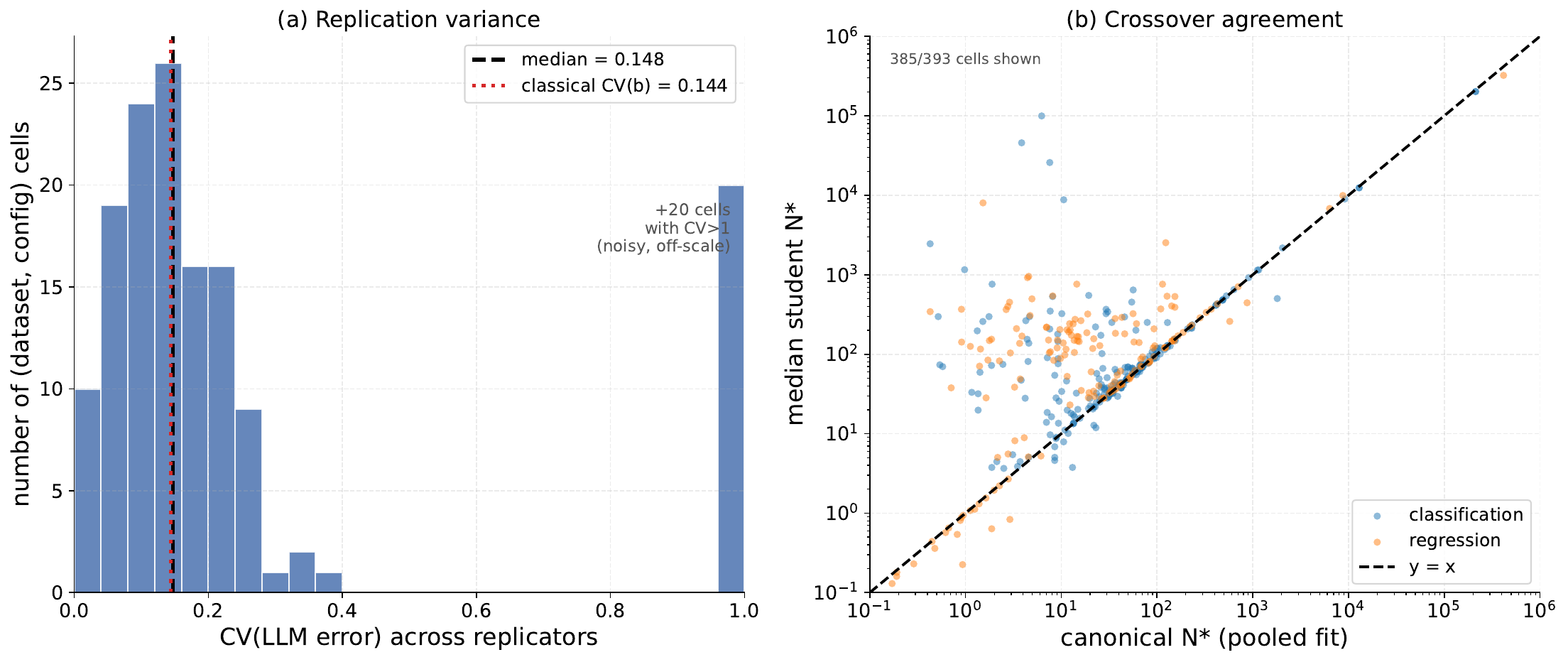}
\caption{(a) Coefficient of variation of LLM error across the $16$--$29$
independent replicators of each (dataset, configuration) cell; the median
($0.148$) is of similar magnitude to the classical cross-student CV ($0.144$),
though the two summarize different quantities. Twenty highly noisy
cells with $\mathrm{CV}>1$ are off-scale. (b) Canonical crossover $N^\star$ from
the pooled fits vs.\ the median of the per-student crossovers; points track the
$y=x$ line.}
\label{fig:variance}
\end{figure}

Because every cell is independently re-implemented by many students, we can ask
two stability questions a single-run LLM evaluation cannot: how much does the
LLM's measured error wander across nominally identical implementations, and does
that wander overturn the train-vs-prompt verdict?

\paragraph{Replication noise is real but bounded.}
Across the $144$ (dataset, configuration) cells, the median coefficient of
variation of LLM error across replicators is \textbf{0.148}
(Figure~\ref{fig:variance}a). As a rough scale reference, the companion study's
cross-student CV of the fitted classical learning-curve exponent is comparable
(\textbf{0.144}); the two summarize different quantities (a bounded error vs.\ a
curve parameter), so we draw no stronger equivalence than this---read loosely, a
$\sim$15\% swing under a nominally fixed protocol is unremarkable for both
prompting and training. The noise is somewhat larger on regression
(CV $0.181$) than classification ($0.134$) and is roughly uniform across the
eight configurations ($0.125$--$0.171$). A tail of $20$ cells has
$\mathrm{CV}>1$, where a near-constant predictor makes the ratio unstable; we
leave these off-scale rather than let them dominate the summary.

\paragraph{The crossover verdict survives the noise.}
Does this variance change the decision? Largely not. Comparing the canonical
pooled-fit crossover against each student's \emph{own submitted} crossover, the
two agree to within a median $|\log_{10}|$ ratio of \textbf{0.118}---well under a
factor of two---across the $393$ comparable cells (Figure~\ref{fig:variance}b).
The qualitative \emph{flags} are concordant: the canonical and student-majority
analyses agree on the \emph{always-better} flag in $614$ of $640$ cells (canonical
flags it in $446$, the student majority in $468$) and on the \emph{never-better}
flag in all $640$ ($39$ each)---that is, they rarely disagree about \emph{whether}
a crossover exists; this is flag concordance, not a claim that the classical model
is always better in $614$ cells. The precise $N^\star$ inherits the replication
uncertainty above, but the decision of \emph{whether} collecting labels is
worthwhile is stable across independent implementations.

\section{Contamination Analysis}
\label{sec:contamination}

The 18 datasets are standard and almost certainly appear in
pretraining~\citep{bordt2024elephants}, so a frozen LLM might \emph{recognize}
rather than \emph{predict} them, inflating its apparent skill and pushing our
crossover rightward. We test this two ways and report a deliberately cautious
conclusion.

\paragraph{Fame does not predict skill (observational).}
We assigned each dataset a $1$--$5$ ``ubiquity'' score for how widely it appears
in public tutorials and repositories, and correlated it with LLM performance.
All associations are null: ubiquity vs.\ best zero-shot error gives Spearman
$\rho=\mathbf{-0.13}$ ($p=0.60$, $n=18$; Figure~\ref{fig:contamination}b), and
ubiquity vs.\ few-shot lift and vs.\ headroom to the classical ceiling are
likewise insignificant ($|\rho|<0.14$, $p>0.58$). Taken alone this would suggest
contamination is not driving results---but a researcher-assigned score is a weak
instrument, so we also intervened directly.

\paragraph{A factorial probe with a trained control (interventional).}
For the same six datasets we move one prompt factor at a time and add a trained
positive control. Anonymizing feature names (keeping the real target description) or
genericizing the task identity (keeping real names) has weak, inconsistent effects
with no significant high$-$low contrast. The telling factor is \emph{value}:
replacing each number with a readable labeled decile (``$d$ (0=low..9=high)''---legible
ordinal information, not opaque indices) collapses the frozen LLM toward chance on
exactly the datasets where it beat chance, while a logistic model trained on the
\emph{same} decile bins keeps its AUROC (at most a $0.09$ drop; the positive control,
Figure~\ref{fig:contamination}a). The LLM's edge is thus \emph{value-dependent}: it
rests on exact feature values a trained model does not need.

\paragraph{But fame is confounded with headroom.}
Readable binning hurts high-ubiquity datasets more than low ones (high$-$low contrast
$+\mathbf{0.25}$, $95\%$ CI $[0.14,0.35]$). Tempting as a contamination signal, this
is confounded: the high-ubiquity datasets are precisely the ones where the LLM beat
chance, so they have the most accuracy to lose under any information-reducing
transform. We therefore cannot attribute the contrast to fame, and---with the
observational null---treat contamination as a \emph{probe}, not a measured effect.
The defensible, now better-supported reading: the frozen LLM leans on recognizable,
exact-valued inputs a genuinely novel table might not supply, which \emph{suggests}---without
proving---that our crossover is a conservative, LLM-favorable estimate
(\S\ref{sec:discussion}).

\section{Discussion}
\label{sec:discussion}

\paragraph{When should you prompt instead of train?}
Rarely, for these tasks and these small, cheap frozen models---essentially only at
the cold start. The crossover sits so far to the left
(\S\ref{sec:crossover}) that a trained model wins using no more than the labels
already on hand in 86\% of cases, and on many datasets is ahead from the smallest
labeled subset we evaluate. The frozen-LLM regime is genuinely useful only at the extreme
cold start---no labels at all, or a one-off prediction where standing up a
training pipeline is not worth it. The moment labeling becomes feasible, the
economics favor training.

\paragraph{You cannot prompt your way across the gap.}
Better prompting does not move the crossover much: the larger model and few-shot
examples shift it only modestly, and serialization format barely at all
(\S\ref{sec:config}). Crucially, in-context examples do not behave like training
data---a zero-to-5-shot jump buys a one-time $-0.06$ in error, but
error-versus-shots follows no power law. The flat LLM line is a ceiling, not a
starting point that more context steadily lifts.

\paragraph{The decision is itself noisy.}
Because independent implementations of the same fixed protocol differ with a
coefficient of variation of $0.148$ in LLM error (\S\ref{sec:variance})---a
spread comparable in scale to the cross-implementer variance of the classical
pipelines---a single team's crossover estimate carries real uncertainty. This argues for reporting the
train-vs-prompt decision as a range, and it cautions against headline LLM
``wins'' that rest on one run of one protocol.

\paragraph{Is the LLM predicting, or recognizing?}
A factorial probe (\S\ref{sec:contamination}) isolates \emph{which} prompt
information the frozen LLM relies on. Anonymizing feature names or dataset identity
has weak, inconsistent effects; the decisive factor is value---replacing exact
numbers with readable deciles collapses the LLM toward chance wherever it beat
chance, while a model trained on the \emph{same} bins keeps its accuracy (a positive
control). The LLM's edge is thus value-dependent. This aligns with dataset fame, but
fame is confounded with baseline accuracy, so we read it as a probe, not proof of
memorization. The takeaway: the frozen LLM leans on recognizable, exact-valued
inputs, which \emph{suggests}, without proving, that its already narrow window would
shrink on a genuinely novel table, making our crossover plausibly conservative.

\paragraph{Cost.}
Even in the small-$N$ region where the LLM is competitive on accuracy, it
carries a recurring per-prediction API cost and latency, whereas a trained
gradient-boosted model has near-zero marginal inference cost
(\S\ref{sec:config}). The accuracy crossover is thus an upper bound on when
prompting is the rational choice once serving cost is included.

\section{Conclusion}
\label{sec:conclusion}

We framed the prompt-vs-train choice as one measurable quantity---the labeled-data
crossover $N^\star$---and estimated it at classroom-replication scale across 18
datasets, eight prompting configurations, and six classical families. For the
regime we study (small, cheap frozen LLMs on business tables) the verdict is
cautionary: even against an oracle choice of the LLM's best configuration, a
trained model wins using no more than the labels already on hand in 86\% of cases
(observed crossover median $\sim$6\%); in-context examples do not close the gap;
and the LLM's skill is value-dependent---it leans on exact, recognizable inputs a
trained model does not need (we do not claim memorization). Asked whether to prompt a small off-the-shelf LLM or train
a gradient-boosted model on a 1{,}000-row table, the evidence says: train it, and
treat the LLM as a cold-start stopgap while the first labels are collected. We
release the de-identified curves, the canonical crossover table, and all analysis
code so practitioners can locate the crossover for their own setting.

\section*{Limitations}
\label{sec:limitations}

\paragraph{Model scope.}
We deliberately study small, inexpensive production models
(\texttt{gpt-4.1-nano}, \texttt{-mini}), the realistic default for a
cost-conscious practitioner. Frontier or task-fine-tuned models would likely
push the crossover $N^\star$ rightward; our claims are about the cheap-LLM
regime, not an upper bound on what any LLM can do. The larger-model
($\texttt{gpt-4.1}$) extension is a within-student paired comparison over $33$ cells
from $11$ students: the bigger frozen model beats nano and mini in ${\sim}85$--$90\%$
of cells yet stays training-free, so it sharpens rather than softens the crossover.
Because \texttt{gpt-4.1} was a self-selected optional run, we still read it as
suggestive, not conclusive.

\paragraph{Test-set size and metric noise.}
Each configuration is scored on up to 100 test rows, so a single cell's
AUROC/RMSE is noisy. We mitigate this with 16--29 independent replicators per cell
and report the resulting variance, but per-dataset point estimates should be read
with that noise in mind.

\paragraph{Contamination.}
The 18 datasets are standard and almost certainly appear in pretraining. Our
probe demonstrates reliance on recognizable feature/target \emph{semantics}---not,
on its own, dataset memorization: the high-vs-low ubiquity contrast is positive
but not significant, and that interval is conditional on six chosen datasets. We
argue this reliance still makes the crossover conservative (a novel, non-semantic
table would offer the LLM less to recognize), but we cannot isolate pretraining
contamination without genuinely novel data, and the probe covers six datasets at
modest sample size.

\paragraph{Replication confound.}
Each student saw only 3 of 18 datasets, so cross-student variance is partially
confounded with dataset-assignment and cohort effects; we therefore frame the
$0.148$ coefficient of variation as protocol/implementation variance, not a
clean random-effects decomposition. The per-student crossover comparison
(\S\ref{sec:variance}) uses students' \emph{own submitted} crossover values, which
carry their individual implementation choices; we read it as a concordance check,
not an independent re-derivation.

\paragraph{Extrapolated crossovers and filter choice.}
$N^\star$ is computed from pooled power-law fits. A minority of cells have
near-flat curves (small exponent $b$), giving unstable or astronomically large
$N^\star$; we flag these and exclude them from headline statistics rather than
over-interpret them. The immediate-win share is stable ($40.6$--$43.1\%$) across
fit-reliability thresholds ($R^2>0.3,0.5,0.7$), and no retained cell lacks a
finite $N^\star$, so the qualitative verdict does not hinge on the filter.

\paragraph{Regression normalization.}
The crossover assumes the classical and LLM regression errors share a
normalization. Both use $\mathrm{RMSE}/\mathrm{std}(y)$ on the original scale, but
the LLM normalizes by the standard deviation of its (up to) 100-row test
\emph{sample} rather than the full test split. We measured this ratio for every
regression dataset (median $1.00$, at most $12\%$ off, scattered above and below
$1$, i.e.\ unbiased) and re-derived all regression crossovers after correcting
for it; the headline observed-crossover median moves only from $5.9\%$ to
$5.2\%$, so the mismatch does not affect our conclusions.

\paragraph{Curve-parameter uncertainty.}
$N^\star$ is most sensitive to the fitted classical parameters $a,b,c$ for the
late- and never-crossing cells. We propagate this uncertainty by bootstrapping the
per-student power-law fits behind each pooled curve (the replication unit,
mean-aggregated) and recomputing $N^\star$: the four-way category of $67$ of $70$
cells is unchanged, and stacking curve-fit uncertainty with the dataset and
test-sample resamplings widens the within-available-data interval only to
$[68,93]\%$ (\S\ref{sec:crossover}). The qualitative verdict therefore does not
hinge on curve-fit uncertainty. The remaining same-distribution check---re-scoring
the best configuration on the full matched test split rather than the $100$-row
sample---leaves the crossover intact and if anything strengthens it
(\S\ref{sec:crossover}), since the LLM's full-split error is slightly higher.

\paragraph{Prompting protocol.}
The LLM is prompted for a hard class label, so its classification AUROC is computed
from $\{0,1\}$ predictions rather than probabilities. Re-scoring ten classification
datasets on the same rows with a probability-eliciting prompt (P$(y{=}1)$,
zero-shot) raises the LLM's AUROC by a median $0.04$ ($8/10$ datasets). A more
competitive LLM raises the error threshold the classical curve must reach, so a
probability-eliciting deployment would push the crossover to modestly larger $N$;
the shift is small and does not overturn the within-available-data verdict, but our
hard-label crossover is best read as a mildly optimistic estimate of how quickly
training overtakes the LLM. The student-contributed few-shot sweep draws in-context exemplar targets
from each implementation's training split; we did not re-verify that regression
exemplar targets were shown on the original (rather than standardized) scale in
every notebook, so the few-shot result is best read as ``in-context examples did
not help as students implemented them'' rather than a fully controlled scaling
study---a reading consistent with its being a negative result.

\paragraph{Domain scope.}
We study English-language prompts on tabular business data. Tables that are
text-heavy, time-series, or in other languages may behave differently, and we do
not evaluate them.

\section*{Acknowledgments}
The measurements analyzed here were produced by the students of STAT 4230/7230 as
independent re-implementations of a fixed course protocol; this study aggregates the
126 usable submissions. I thank them for the effort that makes a classroom-scale
replication possible, and the companion classical scaling-law study for the
authoritative power-law learning-curve fits used to locate the crossover. All student
identifiers are pseudonymized in the released data.

\section*{Data and Code Availability}
The aggregated (pseudonymized) data and the full analysis pipeline are publicly
available at \url{https://github.com/dingkaihua/prompt_or_train} (code under the MIT
license, data under CC BY 4.0). Raw per-student submissions are not released; every
subject key is an opaque pseudonym.

\bibliography{references}

\appendix

\section{Contamination Probe Detail}
\label{app:contamination}

\begin{figure}[h]
\centering
\includegraphics[width=\columnwidth]{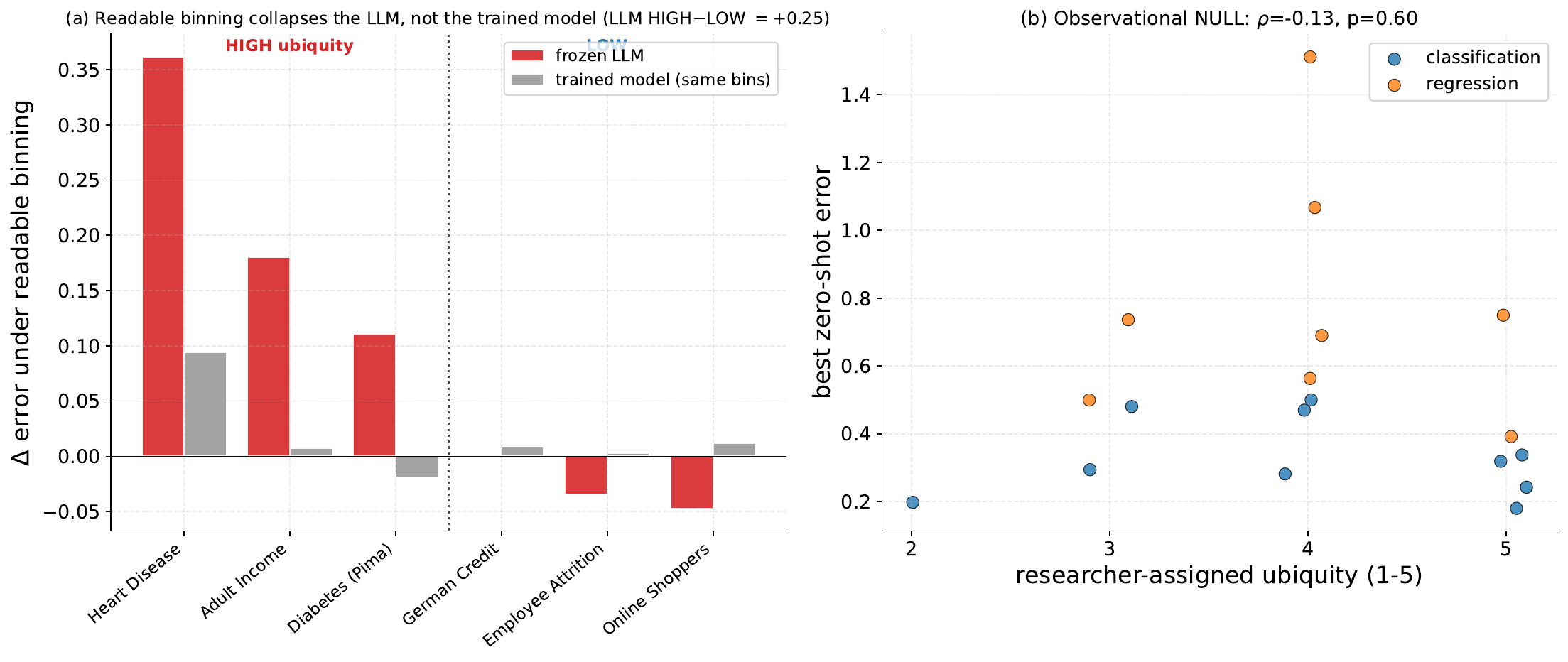}
\caption{Contamination probe (\S\ref{sec:contamination}). (a) Increase in error
when each numeric value is replaced by a readable labeled decile, for the frozen
LLM (red) versus a logistic model trained on the \emph{same} decile bins (grey, the
positive control), across three high- and three low-ubiquity datasets. Readable
binning collapses the LLM toward chance on the high-ubiquity datasets---exactly the
ones where it beat chance---while the trained model is barely affected (LLM
high$-$low contrast $+0.25$, $95\%$ CI $[0.14,0.35]$, but confounded with baseline
accuracy). (b) Observationally, researcher-assigned dataset ubiquity does not
predict the best zero-shot error ($\rho=-0.13$, $p=0.60$).}
\label{fig:contamination}
\end{figure}

\end{document}